\documentclass[letterpaper, 10 pt, conference]{ieeeconf}  % Comment this line out if you need a4paper

\IEEEoverridecommandlockouts    % This command is only needed if 
\usepackage{graphics} % for pdf, bitmapped graphics files
\usepackage{epsfig} % for postscript graphics files
\usepackage{mathptmx} % assumes new font selection scheme installed
\usepackage{times} % assumes new font selection scheme installed
\usepackage{amsmath} % assumes amsmath package installed
\usepackage{amssymb}  % assumes amsmath package installed

\usepackage{amsfonts}
\usepackage{algorithmic}
\usepackage{graphicx}
\usepackage{subcaption}
\usepackage{booktabs}
\usepackage{varwidth}
\usepackage{textcomp}
\usepackage{xcolor}
\usepackage{float}

\usepackage{caption}
\usepackage{comment}
\usepackage{url}
\usepackage{multirow}
\usepackage{multicol}
\title{\LARGE \bf
SwinMTL: A Shared Architecture for Simultaneous Depth Estimation and Semantic Segmentation from Monocular Camera Images}

\author{Pardis Taghavi$^{1}$, Reza Langari$^{1}$ and Gaurav Pandey$^{2}$% <-this % stops a space
\thanks{$^{1}$Pardis Taghavi and Reza Langari
are with the Department of Mechanical Engineering, Texas A\&M University, College Station, TX 77840, USA
{\tt\small \{ptgh, rlangari\}@tamu.edu}
}
\thanks{$^{2}$Gaurav Pandey is with the Department of Engineering Technology \& Industrial Distribution, Texas A\&M University, College Station, TX 77840, USA
{\tt\small gpandey@tamu.edu}}
}

\begin{document}
%\ninept
\maketitle
\thispagestyle{empty}
\pagestyle{empty}

\begin{abstract}
This research paper presents an innovative multi-task learning framework that allows concurrent depth estimation and semantic segmentation using a single camera. The proposed approach is based on a shared encoder-decoder architecture, which integrates various techniques to improve the accuracy of the depth estimation and semantic segmentation task without compromising computational efficiency. Additionally, the paper incorporates an adversarial training component, employing a Wasserstein GAN framework with a critic network, to refine model's predictions. The framework is thoroughly evaluated on two datasets - the outdoor Cityscapes dataset and the indoor NYU Depth V2 dataset - and it outperforms existing state-of-the-art methods in both segmentation and depth estimation tasks. We also conducted ablation studies to analyze the contributions of different components, including pre-training strategies, the inclusion of critics, the use of logarithmic depth scaling, and advanced image augmentations, to provide a better understanding of the proposed framework. The accompanying source code is accessible at \url{https://github.com/PardisTaghavi/SwinMTL}.
\end{abstract}

\section{INTRODUCTION}
Recent advancements in computer vision have enabled various applications ranging from self-driving cars and robot navigation to immersive augmented reality experiences. The key to these advancements lies in scene understanding, where algorithms for semantic segmentation and depth estimation play pivotal roles. Leveraging these algorithms allows for 3D scene reconstruction from camera images alone, reducing the need for expensive sensors such as Lidars, and increasing scalability in real-world scenarios. An additional benefit of obtaining detailed scene understanding from low-cost cameras is that it reduces the need for costly high-definition maps, which may not always be reliable in dynamic environments.

While individual tasks like depth estimation and semantic segmentation have attracted considerable attention and attained impressive accuracy through the evolution of neural network architectures, the concurrent execution of multiple networks on embedded systems poses challenges due to the significant memory footprint and real-time performance constraints. In response, multi-task learning (MTL) has emerged as a solution, seeking to enhance generalization by leveraging domain-specific information within the training signals of interconnected tasks \cite{vandenhende2021multi}.

This paper presents a novel multi-task learning approach utilizing a shared encoder and decoder for concurrent dense semantic segmentation and depth prediction tasks using a single camera. We also integrate adversarial training \cite{arjovsky2017wasserstein} into the training framework that involves training our model alongside a critic network. This results into an iterative refinement process where the critic guides the model towards better predictions for both tasks. The seamless integration of the encoder and decoder in our model significantly reduces the memory footprint of the inference engine, and the adversarial training improves accuracy. Overall our approach yields improved results compared to state-of-the-art methods. Our method is validated on the challenging Cityscapes \cite{cordts2016cityscapes} outdoor dataset and the NYU Depth V2 \cite{silberman2012indoor} indoor dataset demonstrating superior performance in comparison to existing approaches for each task.

\begin{figure}[]
    \centering
        \includegraphics[width=0.52\textwidth]{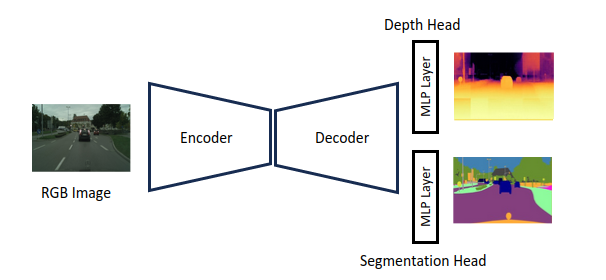}
        
        \caption{Unified Vision: This figure provides a general overview of the seamless integration designed for joint semantic segmentation and depth estimation, ultimately contributing to the creation of a 3D scene.}
    \label{fig:general}
\end{figure}

\section{RELATED WORK}
\subsection{Semantic Segmentation}
The pioneering work in deep learning for semantic segmentation was the introduction of the Fully Convolutional Networks (FCN) architecture by Long et al. \cite{long2015fully}. Despite its effectiveness, FCNs struggle with capturing fine image details due to multiple pooling operations and neglected pixel relationships \cite{chen2020image}. Subsequent works, such as U-net \cite{ronneberger2015u}, addressed these challenges by introducing an encoder-decoder architecture with skip connections to enhance feature propagation. DeepLab \cite{chen2017deeplab} incorporated Atrous Spatial Pyramid Pooling (ASPP) to capture both local and global contextual information for precise segmentation. The recent introduction of attention mechanisms and the development of Vision Transformers (ViTs) \cite{dosovitskiy2020image} has surpassed state-of-the-art convolution-based methods in semantic segmentation \cite{thisanke2023semantic}.

\subsection{Depth Estimation}
Eigen et al \cite{eigen2014depth} pioneered the use of a learning-based approach, introducing a multi-scale convolutional neural network (CNN) coupled with a scale-invariant loss function for monocular depth estimation. Subsequently, several CNN-based methods have tackled depth estimation, treating it as a continuous regression problem based on a single RGB image \cite{LiCRF,HuanOrdinal,KimLeverage}. Some methodologies, exemplified by \cite{fu2018deep} and \cite{diaz2019soft} focused on discretizing continuous depth into intervals, treating depth prediction as a per-pixel classification problem. Another notable approach is found in the work of Bhat et al. \cite{bhat2021adabins}, which combines regression and classification models, formulating depth estimation as a per-pixel classification-regression model. More recently, transformer architectures have made their way into depth estimation networks, aiming to capture global features and enhance prediction accuracy. One notable example is the work of Xie et al. \cite{xie2023revealing}, where transformers were incorporated into depth estimation networks, showcasing their ability to reveal intricate global patterns and improved overall performance.

\subsection{Multi Task Learning}
Multi-task learning (MTL) is a machine learning paradigm where a single model is trained on multiple related tasks \cite{caruana1997multitask}. MTL has gained attention for its capacity to enhance performance on all tasks compared to independent single-task training. This is achieved by leveraging shared information and representations across the tasks, allowing the model to learn more efficiently.
Methods such as \cite{OzenMulti, MarvinMulti}, focus on designing the encoder architecture to learn features that are relevant to multiple tasks. This approach can be effective for tasks with significant shared visual features. However, it may not be optimal for capturing both shared characteristics and task-specific variation features \cite{vandenhende2021multi}. Other works, like \cite{xu2018pad}, propose architectures that predict intermediate auxiliary tasks before using them as additional input for the final task predictions. This approach encourages the model to learn intermediate representations useful for multiple tasks. Beyond architecture design, some studies focused on the optimization process or proposed novel loss functions. For instance, \cite{kendall2018multi} weighs multiple loss functions based on task uncertainty. \cite{chen2018gradnorm} introduced the gradient normalization in deep multitask models to address issues with imbalanced gradients in MTL settings. Additionally, \cite{guo2018dynamic} proposed dynamic task prioritization during training to focus on more challenging tasks. Dynamic prioritization is achieved by adaptively adjusting the weight of each task's loss function based on its perceived difficulty, which is inversely proportional to the model's performance on that task.
 
Xu et al. \cite{xu2018pad} rely on a joint CNN model to fuse information from predicted auxiliary tasks, enhancing accuracy but at the cost of increased model complexity. Hoyer et al. \cite{hoyer2021ways} employ self-supervised depth estimation and transfer learned knowledge to semi-supervised semantic segmentation. This method addresses the challenge of requiring large labeled training data, but it depends on the quality of self-supervised depth estimations and struggles with domain discrepancies. The approach proposed by Liu et al. \cite{liu2019end} tackles the limitations of prior works by employing a single shared network with task-specific attention modules. This combination enables the network to learn shared features across different tasks while enabling each task to focus on its specific feature set through the attention modules. While attention modules can automatically select features, they may not always capture the exact features most relevant to the task. To address this, Ye et al \cite{ye2023taskprompter} introduced explicit task prompts, which are learnable tokens capturing task-specific information. Nevertheless, the inclusion of learnable task prompts introduces significant computational overhead to the network.

The work published by \cite{xu2018pad, hoyer2021ways, liu2019end} is closest to ours as they apply MTL to joint estimation of depth and semantic segmentation. Similar to Liu et al \cite{liu2019end}, we employ an end-to-end training approach for joint depth estimation and semantic segmentation. However, unlike their use of task-specific attention modules, our method directly shares features across tasks, further simplifying the model. Additionally, the incorporation of an adversarial training framework maintains high accuracy. This design choice is crafted to overcome limitations observed in related works, where the introduction of extra network components resulted in increased complexity. Our primary objective is to showcase enhanced accuracy through the seamless optimization of weights for both depth regression and pixel classification. By sidestepping the challenges associated with auxiliary tasks such as task-specific attention modules and complexities, our approach presents a more practical and effective solution. %Moreover, we leverage a Stereo Matching neural network to enhance the reliability of depth annotations for the Cityscapes dataset. %
% We evaluate the performance of our integrated network on both the Cityscapes and NYU Depth V2 datasets, showcasing our improved outcomes across diverse tasks and datasets.

\begin{figure*}
    \centering
    \includegraphics[width=1\linewidth]{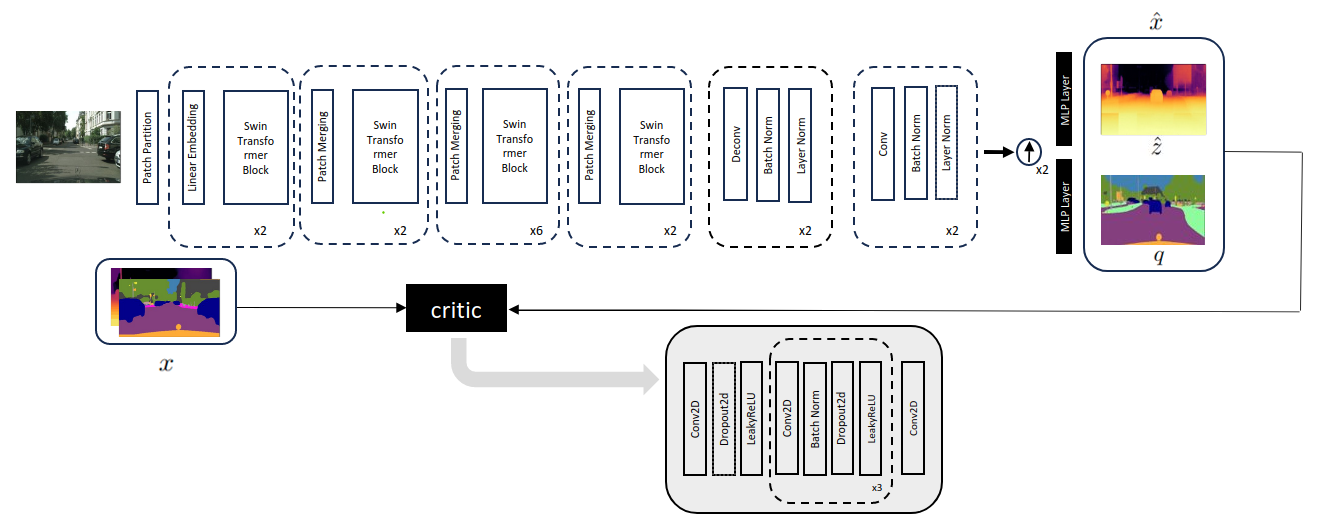}
    \caption{Network Architecture: Figure shows the detailed network architecture tailored for the precise prediction of both depth maps and segmentation maps. Notably, the network employs two critic discriminators during training, enhancing its ability to discern and refine the joint output representing depth and segmentation information.}
    \label{fig:architecture}
\end{figure*}

\section{METHOD}

\subsection{Model Architecture}
In our proposed model, we adopt the vision transformer introduced by Liu et al. \cite{liu2021swin} as our backbone. The inherent hierarchical structure and attention mechanisms of transformers make them well-suited for capturing complex spatial relationships, which are crucial for tasks such as depth estimation and semantic segmentation. The Swin Transformer\cite{liu2021swin}, in particular, demonstrates superior performance owing to its additional shifted windows approach in the backbone. In our method, we design a multi-task learning framework to simultaneously address both depth estimation and semantic segmentation task. Notably, we deviate from the prevalent trend of employing intricate architectures for multi-task scenarios, opting for a consistent encoder and decoder for both tasks. This departure allows us to maintain shared components, simplifying the overall architecture and rendering it suitable for embedded systems. Additionally, we incorporate a generative adversarial training approach that enhances the overall accuracy. While previous studies leveraging GANs typically utilize binary classification discriminators in similar settings, we opt for the Wasserstein GAN (WGAN) \cite{arjovsky2017wasserstein} to measure the similarity distance between predictions and ground truth distributions. Our approach showcases superior performance on both tasks when compared to training individual networks for each task with the same computational cost.

To enable individualized predictions for each task, we introduce two task-specific Multi-Layer Perceptron (MLP) layers at the end of the decoder. One MLP layer is dedicated to depth prediction, while the other is tailored for semantic segmentation. This streamlined and shared architecture contributes to the model's memory efficiency and ensures high-quality task-specific predictions. The shared components allow for seamless integration of information, promoting generalization across tasks, while the task-specific MLP layers enable the model to fine-tune its predictions for depth estimation and semantic segmentation independently. Next we describe the loss functions used to train the proposed network.

\textbf{Depth Estimation Loss}: We utilize the pixel-wise scale-invariant loss proposed by \cite{eigen2014depth} for depth estimation:

\begin{equation}
    L_{depth}(\hat{z}, z_{gt}) = \frac{1}{N} \sum_{i,j} (e_{i,j}^2) - \frac{\alpha}{N^2}(\sum_{i,j} e_{i,j})^2 
    \label{eq:depth-loss}
\end{equation}
where
\begin{equation}
    e_{i,j} = \log \hat{z}^{i,j} - \log z_{gt}^ {i,j}
    \label{eq:depth-log-loss}
\end{equation}
    
Here $z^{i,j}$ is the depth at pixel $(i,j)$. The loss is designed to be invariant to global scale changes in the predicted depth map. The Logarithmic term \eqref{eq:depth-log-loss} reduces the impact of large errors, making it robust to outliers and more accurate as compared to vanilla $L1$ or $L2$ loss.

\textbf{Segmentation Loss}: We adopt the standard cross-entropy loss for segmentation, effectively guiding the model towards accurate pixel-level classification:
\begin{equation}
    L_{seg}(p, q) = -\sum_i p_i \log(q_i)
    \label{eq:seg-loss}
\end{equation}

In equation \eqref{eq:seg-loss}, $p$ represents the ground truth label of each pixel and $q$ is the predicted probability distribution obtained from the log-softmax function. For a given class $i$, \( q_i \) is calculated using the log-softmax function:

\begin{equation}
     q_i = \log\left(\frac{e^{x_i}}{\sum_j e^{x_j}}\right)
     \label{eq:seg-logit}
\end{equation}

In equation \eqref{eq:seg-logit}, $x_i$ represents the logit corresponding to class $i$, and $\sum_j e^{x_j}$
represents the sum of logits for all classes. This formulation ensures a robust and normalized representation of predicted probabilities for accurate segmentation loss computation.

\textbf{Adversarial Loss}: We integrate the loss function proposed in \cite{gulrajani2017improved} for the generative adversarial networks, incorporating a gradient penalty (GP) regularization term to enforce Lipschitz continuity on the critic ($c$), promoting stability and smoothness in training. This technique prevents the critic's gradients from becoming excessively large, which can lead to training instability and convergence issues.

\begin{equation}
    L_{GAN}(\hat{x}, x) = \mathbb{E} (c(x)) - \mathbb{E} (c(\hat{x})) + \lambda * GP  
    \label{eq:adv-loss1}
\end{equation}

\begin{equation}
     GP = \mathbb{E} [(\|{\nabla_{\Tilde{x}}c(\Tilde{x})\|_{2}} -1)^2]
     \label{eq:adv-loss2}
\end{equation}

\begin{equation}
    \Tilde{x}= \epsilon x + (1-\epsilon)\hat{x}
    \label{eq:adv-loss3}
\end{equation}

In equation \eqref{eq:adv-loss1}, $L_{GAN}(\hat{x},x)$ represents the adversarial network loss, where $\hat{x}$ is the generated sample and $x$ is the real sample(Fig. \ref{fig:architecture}). The gradient penalty term GP \eqref{eq:adv-loss2} enforces smoothness by penalizing deviations from Lipschitz continuity. The parameter $\lambda$ controls the strength of the regularization. Equation \eqref{eq:adv-loss3} introduces $\Tilde{x}$, a linear interpolation between real and generated samples, facilitating gradient computation for the penalty term.

\textbf{Total Loss}: Total loss is a balanced combination of depth estimation, segmentation, and adversarial network losses, with $\alpha$ and $\beta$ controlling the influence of each loss function:

\begin{equation}
     L_{\text{tot}} = \alpha L_{\text{depth}} + (1-\alpha) L_{\text{seg}} + \beta L_{\text{GAN}}   
     \label{eq:total-loss}
\end{equation}

Adversarial learning is a pivotal component of our approach. Inspired by Generative Adversarial Networks (GANs) \cite{goodfellow2014generative}, our approach establishes a dynamic interplay between two competing networks: the generator ($g$), represented by SwinMTL, and the discriminator (referred to as a critic in the context of \cite{arjovsky2017wasserstein}). This interplay engages these two networks in a min-max competition, as shown in \eqref{eq:valueFunction}, where predictions are refined towards ground truth labels, enhancing overall accuracy.

In implementing adversarial training, we introduce a critic network during the training process to refine the output of the generator. Unlike previous approaches that employ discriminators through binary classification, we leverage the Wasserstein GAN framework with Earth Mover's Distance (EMD) \cite{arjovsky2017wasserstein, gulrajani2017improved}. EMD quantifies dissimilarity between two distributions, namely predictions($\hat{x}$) vs ground truth labels($x$), addressing issues such as mode collapse and preventing the generator from outperforming the discriminator in the early epochs.

The critic in this approach receives as input the concatenated depth and segmentation maps of both ground truths and predictions. It maximizes the value function defined in \eqref{eq:valueFunction}, aiming to differentiate between the ground truth and the predicted data by outputting a probability ranging from zero to one. Conversely, the generator endeavors to minimize the value function to reduce the EDM distance between these datasets. This adjustment allows the critic to thoroughly assess the alignment between generated and authentic data across various dimensions. The detailed architecture of the critic is delineated in Fig. \ref{fig:architecture}.

\begin{equation}
\min_{g}\max_{c}\mathbb{E}_{x\sim p_{\text{data}}} [c(x)] - \mathbb{E}_{x\sim p_{\text{g}}} [c(g(x))] 
\label{eq:valueFunction} 
\end{equation}

\section{Experiments}
In this section, we present the details of our experimental setup, showcasing results on the Cityscapes and NYU Depth V2 datasets. Additionally, we conduct ablation studies to gain insights into the individual contributions of key components in our approach.

\begin{center}
\begin{figure}
\centering
\includegraphics[width=0.5\textwidth]{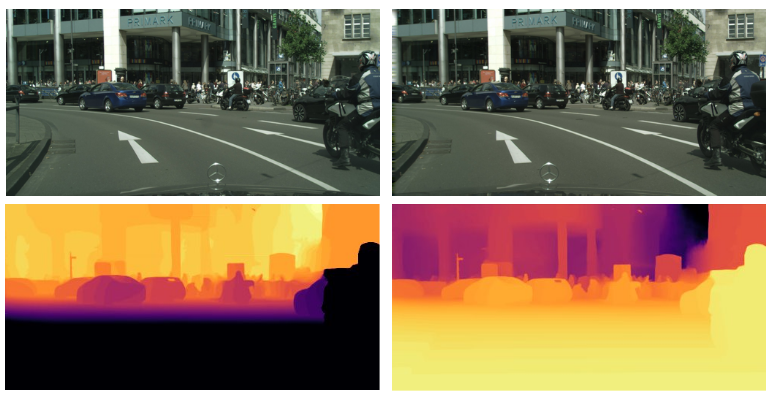}
\caption{The first row shows a stereo pair from the Cityscapes dataset. The second row illustrates the dense disparity map (bottom right) and the dense depth map (bottom left), both generated using the advanced Cascaded Recurrent Stereo Matching Network (CREStereo) \cite{li2022practical}}
\label{fig:stereo}
\end{figure}
\end{center}

\subsection{Experimental Setup}
Cityscapes is an outdoor dataset capturing urban street scenes, Cityscapes provides high-quality pixel-level annotations for 30 classes grouped into 8 categories (19 classes for training). The dataset comprises approximately 5000 finely annotated images, distributed into 2975, 500, and 1525 images for training, validation, and testing, respectively. Spanning 50 different cities in Germany, the dataset's resolution is 2048x1024. The Cityscapes dataset does not have dense depth annotation but includes stereo images.  We leverage these stereo images to generate the dense disparity map using the Cascaded Recurrent Stereo Matching Network (CREStereo) \cite{li2022practical} and, accordingly, the dense depth map (Fig. \ref{fig:stereo}) required for training our network. NYU Depth V2 dataset consists of 1449 RGB-D images of indoor scenes, each paired with dense labeled depth images and semantic labels across 40 categories. Each image is of size 640x480, distributed into 795 training and 654 testing images. 

Inspired by \cite{li2023openmixup} we integrated a modified version of MixUp augmentation into our training pipeline to increase the robustness and generalization capabilities of our model. we randomly select a fixed number ($N$) of square patches from each input image. These patches vary in size, ranging from $1/8$ to $1/2$ of the image height ($h$). The patches are then shuffled, disrupting the original spatial relationships within the image. The shuffled patches are displaced within images of each batch to create new augmented images in each epoch. This process effectively blends visual information from multiple images, generating a greater diversity of training samples. MixUp Augmentation proves particularly beneficial in the NYU Depth V2 dataset, mitigating challenges arising from its limited number of images.

\begin{table*}
\begin{center}
\caption{Quantitative comparison of single-task depth prediction methods on Cityscapes test set (starred methods are self-supervised)}
\begin{tabular}{|c|c|ccccccc|}
\hline
model  & Resolution&  AbsRel $\downarrow$& SqRel $\downarrow$& RMSE $\downarrow$ & RMSE log $\downarrow$  & $\delta<1.25$ $\uparrow$ & $\delta<1.25^2$ $\uparrow$ & $\delta<1.25^3$ $\uparrow$\\
\hline
GLNet\cite{lee2021learning} & 1024x2048 & $0.111$ & -& $6.437$ & $0.182$  & $0.868$ & $0.961$ & $0.983$ \\
Pilzer et al.\cite{pilzer2018unsupervised}* &  512x256 & $0.438$ & $5.713$ & $5.745$ & $0.400$ & $0.711$ & $0.877$ & $0.940$ \\
Struct2Depth\cite{casser2019unsupervised}* &  416x128 & $0.145$& $1.737$ & $7.280$ & $0.205$ & $0.813$ & $0.942$ & $0.976$ \\
%Monodepth2\cite{}* &  416x128 & $0.129$ & $1.569$ & $6.876$ & $0.187$ & $0.849 $ & $0.957$ & $0.983$ \\
Li et al.\cite{li2021unsupervised}* &  416x128 & $0.119$ & $1.290$ & $6.980$ & $0.190$ & $0.846$ & $0.952$ & $0.982$ \\
DynamicDepth\cite{feng2022disentangling}* &  416x128 & $0.103$ & $1.000$ & $5.867$ & $0.157$ & $0.895$ & $0.974$ & $0.991$ \\
Lee et al.\cite{lee2021attentive}* &  832x256 & $0.116$ & $1.213$ & $6.695$ & $0.186$ & $0.852$ & $0.951$ & $0.982$ \\
InstaDM\cite{lee2021learning}* &  832x256 & $0.111$ & $1.158$& $6.437$ & $0.182$ & $0.868$ & $0.961$ & $0.983$ \\
\hline\hline
SwinMTL & 512x1024 & \textbf{0.089} & \textbf{1.051} & \textbf{5.481} & \textbf{0.139} & \textbf{0.921} & \textbf{0.976} & \textbf{0.990}  \\
\hline
\end{tabular}
\label{table:1}
\end{center}
\end{table*}

\begin{figure}[h]   
    \centering
    \includegraphics[width=1\linewidth]{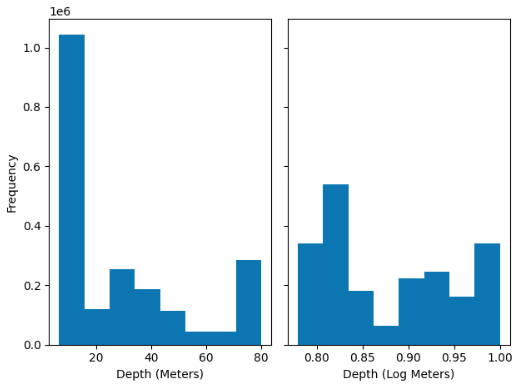}
    \caption{Histogram of depth values of an instance in the Cityscapes dataset, visually highlighting the dominance of values within the $0-10$ meter range. The distribution's peak and long tail motivate the logarithmic transformation.}
    \label{fig:depthProb}
\end{figure}

To underscore the importance of the near-range region for depth estimation in Cityscapes dataset, we transform the depth from linear space to logarithmic space \cite{saxena2023zeroshot} for the critic network. As illustrated in Fig. \ref{fig:depthProb}, a significant concentration of pixels falls within the 0-10 meter range, constituting over $80\%$ of pixels in each image. In order to effectively capture this distribution and assign greater significance to this critical zone, we employ the logarithmic depth parameterization proposed in \cite{saxena2023zeroshot}. This transformation aligns model learning with the natural distribution of depth values, resulting in substantial improvements in depth estimation accuracy for the Cityscapes dataset. Here, $d_{metric}$ denotes the metric depth, while $d_{min}$ and $d_{max}$ represent the minimum and maximum depth values respectively.

\begin{equation}
  d_{log}= \frac{\log (d_{metric}/d_{min})}{\log(d_{max}/d_{min})}  
  \label{eq:log-depth}
\end{equation}

%\textbf{Evaluation Metrices} We report results using standard metrics employed by state-of-the-art methods. For depth estimation, we utilize the square root of the Scale Invariant Logarithmic error (SILog), Relative Squared error (Sq Rel), Relative Absolute Error (Abs Rel), Root Mean Squared error (RMS), and threshold accuracy (percentage of pixels where the estimated depth falls within a certain threshold of the ground truth value) \cite{eigen2014depth}. For semantic segmentation, we employ the mean Intersection-over-Union (mIoU) metric \cite{everingham2010pascal}.

SwinMTL offers flexibility in choosing different backbones for its model; however, our reported results are based on the SwinV2-B model, pretrained with Masked Image Modeling (MIM) \cite{xie2023revealing}. We employ the AdamW optimizer \cite{loshchilov2017decoupled} with a base learning rate of 5e-4 and weight decay of 5e-2. The learning rate follows a polynomial strategy with a decay factor of 0.9. Additionally, we incorporate random adjustments in brightness, contrast, gamma correction, hue saturation value, and horizontal flip, aligning with established methodologies in prior research. For direct comparison with other approaches, we set the maximum depth range to $80m$ for the Cityscapes dataset and $10m$ for the NYU Depth V2 dataset. We implemented our method in PyTorch 2.0.1 (Python 3.8) with CUDA 11.8, utilizing two NVIDIA GeForce RTX 3090 GPUs.

\subsection{Results and Comparisons}
\begin{figure*}
    \centering
    \includegraphics[width=1\linewidth]{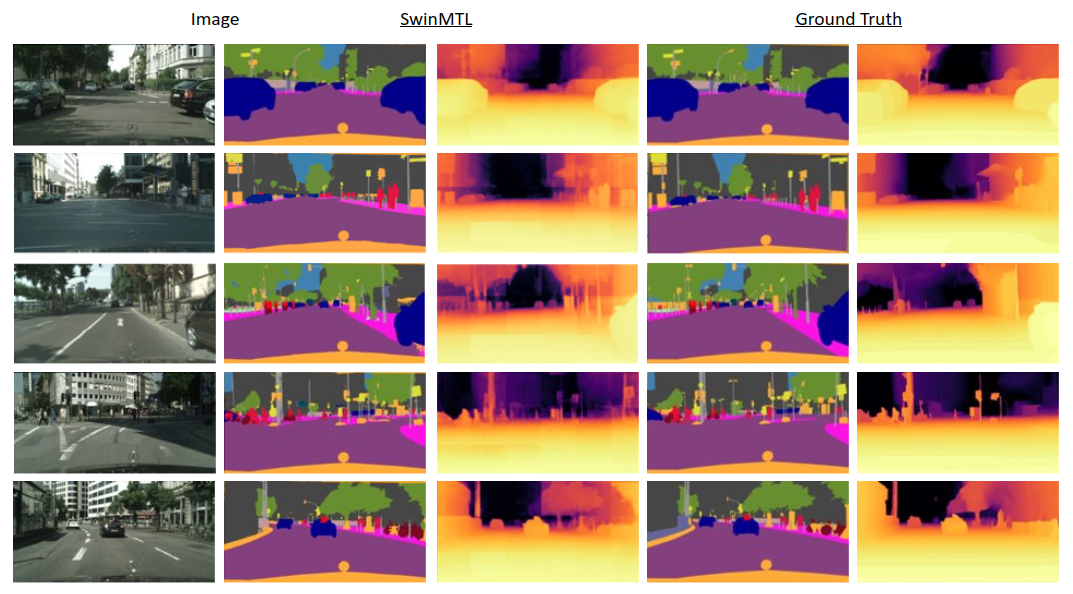}
    \caption{Exploring SwinMTL's Qualitative Results on the Cityscapes Dataset}
    \label{fig:qual}
\end{figure*}

We conduct a thorough performance comparison of SwinMTL against state-of-the-art models on the Cityscapes and NYU Depth V2 datasets for dense prediction tasks. The results for the Cityscapes dataset are compared with other depth estimation methods in Table \ref{table:1} and with other MTL networks in Table \ref{table:tabMLT}. Table \ref{table:nyu2} showcases the performance of NYU Depth V2 dataset compared to other pioneering MTL networks. We ensure a fair comparison by using the best-reported numbers from prior works. As demonstrated by the tables for both Cityscapes and the NYU Depth V2 dataset, SwinMTL consistently outperforms or shows comparable performance with enhanced computational efficiency. For additional insights, Fig. \ref{fig:qual} presents qualitative results on the Cityscapes dataset.

%CITYSCAPES

\begin{table}[]
\caption{Quantitative Performance Comparison of Our MTL Framework versus SOTA Methods on Cityscapes Test Set on The Full Resolution (1024x2048)}
\begin{center}
\begin{tabular}{|c|cc|}
\hline
model &  RMSE $\downarrow$ & mIoU $\uparrow$\\
\hline
Hybridnet\cite{escohybrid} &  12.090 & 66.61\\
MGMNet\cite{Schon_2021_ICCV} & 8.300 & 55.70\\
MTL\cite{vandenhende2021multi} &  6.797 & 70.43\\
PAD-Net\cite{xu2018pad} & 6.777 &  70.23\\
3-ways\cite{hoyer2021ways} &  6.528 & 75.00\\
\cite{lopes2022crosstask} &  6.649 & 74.95\\
\hline\hline
SwinMTL & \textbf{6.352} & \textbf{76.41}\\
\hline
\end{tabular}
\label{table:tabMLT}
\end{center}
\end{table}

%NYU
\begin{table}[]
\caption{Quantitative Performance Comparison of SwinMTL Framework versus SOTA Methods on NYU Test Set}
\begin{center}
\begin{tabular}{|c|cc|}
\hline
model &  RMSE $\downarrow$ & mIoU $\uparrow$\\
\hline
Cross-Stitch\cite{misraStitch}         & $0.6290$ & $36.34$\\
MTL\cite{vandenhende2021multi}         & $0.6380$& $39.44 $\\
PAD-Net\cite{xu2018pad}                & $0.6270$ &  $36.61$ \\
3-ways\cite{hoyer2021ways}             & $0.622$ &  $39.47$ \\
\cite{lopes2022crosstask}              & $0.6040$ &  $38.93$ \\
MTAN\cite{liu2019end} &$0.5906$ & $40.01$\\
MTI-Net\cite{VandenhendeMTI}           & $0.5365$ &  $45.97$ \\
ATRC\cite{BruggemannATRC}              & $0.5363 $  & $ 46.33$ \\
InvPT\cite{ye2022invpt}                & $0.5183$ & $53.56$\\
TaskPrompter\cite{ye2023taskprompter}  & $\textbf{0.5152}$ &  $55.30$\\
\hline\hline
SwinMTL        & $0.5179$ & $\textbf{58.14}$ \\
\hline
\end{tabular}
\label{table:nyu2}
\end{center}
\end{table}

\subsection{Ablation Studies}
To efficiently assess the individual contributions of various components within our proposed framework, we conducted ablation studies on lower resolution images (256x128) from the Cityscape dataset for accelerated computation. These findings provide valuable insights into the effectiveness of each module while maintaining computational feasibility.

\textbf{Comparison of Single and Multi-task Networks:} We conducted a comparative analysis of single-task and multi-task networks using a Swin-Base backbone. As summarized in Table \ref{ablation}, the multi-task network achieved superior evaluation metrics for both depth estimation and semantic segmentation tasks, despite a marginal increase of only $3\%$ in parameter count. This evidence showcases the potential of multi-task learning to enhance model performance without compromising efficiency.

\begin{table}[]
\caption{Ablation Studies on the Cityscapes dataset}
\begin{center}
\begin{tabular}{|c|c|cc|}
\hline
Ablation  & Model & AbsRel$\downarrow$ & mIoU$\uparrow$\\
\hline

\multirow{3}{9em}{Single-Task vs. Multi-Task Network Performance } & only Depth & $0.110$ & -\\
 & only Seg & - & $42.82$\\
& SwinMTL &  $0.104$ & $50.35$\\
\hline
\multirow{3}{9em}{Impact of Different Pretrained Networks} & w/o pretrained & $0.617$ & $17.24$\\
& w normal pretrained & $0.432$ & $25.73$\\
& w MIM pretrained &  $0.104$ & $50.35$\\
\hline
\multirow{3}{9em}{Impact of Critic Integration} & w/o Critic & $0.234$ & $50.15$\\
& w one Critic & $0.104$ & $50.35$\\
& w two Critics & $0.105$& $50.30$\\
\hline
\multirow{2}{9em}{Linear vs. Logarithmic} & Linear Space & $0.152$ & -\\
& Log Space & $0.104$ & - \\
\hline
\end{tabular}
\label{ablation}
\end{center}
\end{table}

%\begin{table}[]
%\caption{Comparison of Single-Task and Multi-Task Network Performance}
%\begin{center}
%\begin{tabular}{|c|ccc|}
%\hline
%model & N. of parameters & AbsRel$\downarrow$ & mIoU$\uparrow$\\
%\hline
%only Depth & $87.21M$ & $0.110$ & -\\
%only Seg & $87.24M$ & - & $42.82$\\
%SwinMTL & $87.38M$ & $0.104$ & $50.35$\\
%\hline
%\end{tabular}
%\label{singlemulti}
%\end{center}
%\end{table}

\textbf{Performance with Different Pretrained Networks:} Understanding the influence of pretraining variations on downstream task performance is crucial for model optimization. In this ablation study (Table \ref{ablation}), we evaluate our model's performance with diverse pretrained networks: a standard Swin-V2 model pretrained on ImageNet-22K and a Swin-V2 pretrained with masked image modeling (MIM) \cite{xie2023revealing}. By comparing results, we can see the effectiveness of MIM pretraining on the accuracy of depth and segmentation predictions which is also aligned with results shown in \cite{xie2023revealing}.

%\begin{table}[]
%\caption{Impact of Different Pretrained Networks on Performance}
%\begin{center}
%\begin{tabular}{|c|cc|}
%\hline
%model &  AbsRel$\downarrow$ & mIoU$\uparrow$\\
%\hline
%w/o pretrained & $0.617$ & $17.24$\\
%w normal pretrained & $0.432$ & $25.73$\\
%w MIM pretrained &  $0.104$ & $50.35$\\
%\hline
%\end{tabular}
%\label{MIM}
%\end{center}
%\end{table}

\textbf{Critic Addition Performance:} The integration of a critic into the network architecture led to performance gains, as evidenced in Table \ref{ablation}. Notably, incorporating a single critic during training, without increasing testing parameters, produced the most significant improvements in depth estimation and semantic segmentation accuracy. This addition highlights the potential for critics to refine model learning without sacrificing computational overhead.

%\begin{table}[]
%\caption{Impact of Critic Integration on Performance}
%\begin{center}
%\begin{tabular}{|c|cc|}
%\hline
%model &  AbsRel$\downarrow$ & mIoU$\uparrow$\\
%\hline
%w/o Critic & $0.234$ & $50.15$\\
%w one Critic & $0.104$ & $50.35$\\
%w two Critics & $0.105$& $50.30$\\
%\hline
%\end{tabular}
%\label{critic}
%\end{center}
%\end{table}

\textbf{Integrating Logarithmic Depth Scaled for Enhanced Accuracy:} Fig.\ref{fig:depthProb} reveals a major concentration of depth values within the 0-10 meter range, accounting for over $80\%$ of the pixels in our dataset. To effectively model this distribution and prioritize precision in this critical zone, we adopt the logarithmic depth parameterization proposed in \cite{saxena2023zeroshot}. This transformation aligns model learning with the natural distribution of depth values, leading to substantial improvements in depth estimation accuracy, as evidenced in Table \ref{ablation}.

%\begin{table}[]
%\caption{Quantitative comparison of depth estimation metrics between linear and logarithmic depth scales, demonstrating the clear performance gains achieved with the logarithmic approach.}
%\begin{center}
%\begin{tabular}{|c|c|}
%\hline
%Model &  AbsRel \\
%\hline
%Linear Space & $0.152$\\
%Log Space & $0.104$ \\
%\hline
%\end{tabular}
%\label{table:log}
%\end{center}
%\end{table}

\textbf{Computation Footprint:} Lastly, table \ref{table:computation} showcases a comparison against state-of-the-art methods, revealing our model's ability to achieve high accuracy with significantly fewer parameters and lower computational costs.

\begin{table}[]
\caption{The parameter count comparison across different methods, emphasizing our model's lean architecture.}
\begin{center}
\begin{tabular}{|c|cc|}
\hline
Models & FLOPs &  N. of Parameters \\
\hline
TaskPrompter\cite{ye2023taskprompter} & $416$G & $373$M\\
ATRC w/ ViT-B\cite{BruggemannATRC}& $260$G    & $119$M\\
PAD-Net w/ ViT-B\cite{xu2018pad} & $166$G & $109$M\\
SwinMTL w/ Swin-B & \underline{$65$G} & \underline{$87.38$M}\\
\hline
\end{tabular}
\label{table:computation}
\end{center}
\end{table}

\section{CONCLUSIONS}

In conclusion, this paper proposed a streamlined yet effective multi-task learning framework for joint depth estimation and semantic segmentation. We demonstrate the effectiveness of our method on the Cityscapes and NYU Depth V2 datasets, consistently outperforming leading-edge methods in both depth estimation and semantic segmentation. The contributions of our work encompass the introduction of a multi-task learning design that considers computational efficiency along with accuracy, the integration of a critic network to refine dense predictions. Furthermore, our work studies the impact of various components, including pretraining strategies, the incorporation of critics, the utilization of logarithmic depth scaling, and the inclusion of MixUp augmentation through extensive ablation studies. Future work could explore further optimizations on joint weight training procedures to deepen our understanding of the intricate relationship between depth and segmentation. Additionally, extending the application of our framework to diverse datasets and varied environmental conditions would contribute to its robust generalization capabilities.

\addtolength{\textheight}{-1cm}   % This command serves to balance the column lengths
                                  % on the last page of the document manually. It shortens
                                  % the textheight of the last page by a suitable amount.
                                  % This command does not take effect until the next page
                                  % so it should come on the page before the last. Make
                                  % sure that you do not shorten the textheight too much.

\bibliographystyle{IEEEbib}
\bibliography{sample}

\end{document}